\documentclass[letterpaper, 10 pt, conference]{ieeeconf}  

\IEEEoverridecommandlockouts

\usepackage{graphicx} 
\usepackage{amsmath,amssymb}
\usepackage{color}
\usepackage[table,xcdraw]{xcolor}  
\usepackage{array}  
\usepackage{multirow}
\usepackage{tabularx} 
\usepackage{float}
\usepackage{algorithm}
\usepackage{algpseudocode}
\usepackage{caption}
\usepackage{subcaption}
\usepackage{soul}
\usepackage{wrapfig}
\usepackage{hyperref}

\title{\LARGE \bf
MoistureMapper: An Autonomous Mobile Robot for \\High-Resolution Soil Moisture Mapping at Scale
}
\author{Nathaniel Rose$^1$, Hannah Chuang$^1$, \\Manuel A Andrade-Rodriguez$^1$, Rishi Parashar$^2$, Dani Or$^1$, Parikshit Maini$^1$
\thanks{$^{1}$University of Nevada Reno, USA $^{2}$Desert Research Institute, Reno, Nevada, USA. This work is supported by the NASA EPSCoR Program and NVSGC as part of the project titled “ARC: Prospecting and Pre-Colonization of the Moon and Mars using Autonomous Robots with Human-in-the-Loop.”\vspace{1mm}
\copyright 2025 IEEE. Personal use of this material is permitted. Permission from IEEE must be obtained for all other uses, in any current or future media, including reprinting/republishing this material for advertising or promotional purposes, creating new collective works, for resale or redistribution to servers or lists, or reuse of any copyrighted component of this work in other works.%
}%
}

\begin{document}

\maketitle

\begin{abstract}    
Soil moisture is a quantity of interest in many application areas including agriculture and climate modeling. Existing methods are not suitable for scale applications due to large deployment costs in high-resolution sensing applications such as for variable irrigation. In this work, we design, build and field deploy an autonomous mobile robot, MoistureMapper, for soil moisture sensing. The robot is equipped with Time Domain Reflectometry (TDR) sensors and a direct push drill mechanism for deploying the sensor to measure volumetric water content in the soil. Additionally, we implement and evaluate multiple adaptive sampling strategies based on a Gaussian Process based modeling to build a spatial mapping of moisture distribution in the soil. We present results from large scale computational simulations and proof-of-concept deployment on the field. The adaptive sampling approach outperforms a greedy benchmark approach and results in up to 30\% reduction in travel distance and 5\% reduction in variance in the reconstructed moisture maps. Link to video showing field experiments: \url{https://youtu.be/S4bJ4tRzObg}

\end{abstract}

\section{Introduction}
Soil moisture plays a key role in agricultural productivity, water management, and climate modeling, with its importance underscored by its designation as one of the top 50 Essential Climate Variables (ECVs) by the Global Climate Observing System (GCOS) \cite{GCOS_2025}. In the context of agriculture, the availability of high-resolution soil moisture data can significantly improve the efficiency of irrigation management systems \cite{irigation}. While extensive efforts have been made to develop technologies for measuring soil moisture, existing methods exhibit critical limitations that hinder their practical utility for large-scale, high-resolution applications like variable rate irrigation. Remote sensing technologies such as satellite-based microwave sensors, though capable of penetrating cloud cover and providing wide coverage, are affected by vegetation attenuation, limiting their effectiveness in areas with dense crop cover. Thermal and optical remote sensing approaches \cite{schmugge1980survey} rely on surface temperature and albedo changes but are sensitive to time of day, cloud cover, and climatic variations, rendering them unreliable under fluctuating environmental conditions. Moreover, the coarse spatial resolution of these methods fails to capture the fine-scale heterogeneity of soil moisture that is crucial for precision irrigation. EM-based methods \cite{davis1977electromagnetic}, including Time Domain Reflectometry (TDR) sensors, capacitance sensors and neutron-based techniques, provide high accuracy measurements at fixed locations. Deploying a dense network of these sensors to achieve comprehensive field coverage is cost-prohibitive, especially for larger agricultural fields. These challenges highlight the need for innovative solutions to address the gaps in soil moisture mapping.

\begin{wrapfigure}{r}{0.5\columnwidth}
    \vspace{-3mm}
    \centering
    \includegraphics[width=0.5\columnwidth]{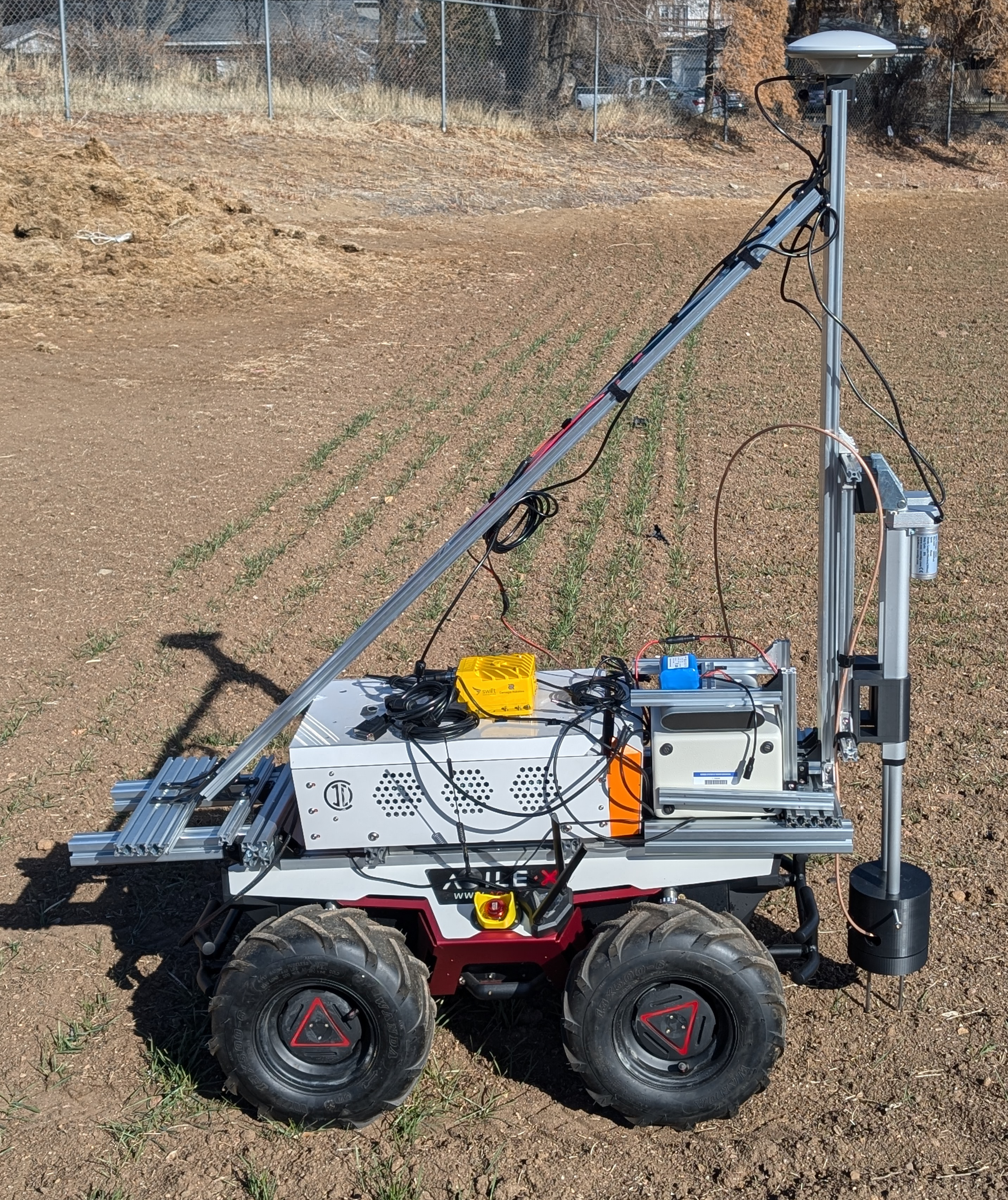}
    \caption{ MoistureMapper: soil moisture sensing robot deployed in the field.}
    \vspace{-3mm}
    \label{fig:MM_Field_Close}
\end{wrapfigure}

To address these issues, in this work we propose to use a wheeled mobile robot equipped with EM-based TDR sensors for soil moisture sensing on agricultural fields (see Fig. \ref{fig:MM_Field_Close}). 
TDR sensors are multi-pronged buriable sensors that estimate the dielectric constant for the soil based on the amount of time it takes for an electric pulse to travel through the probes of the sensor when buried in the soil \cite{Jones2003}. While these sensors have been in use for measuring soil moisture for over two decades now, they need to be installed under the soil at the desired depth (typically by excavating the soil and burying the sensor) and connected to a data logger using wired connections. For high-resolution moisture mapping, a multitude of these sensors can be deployed. At scale, this leads to high costs for sensor acquisition, deployment and upkeep of the network of data loggers and wired connections, restricting its deployment on the farms. We design a sensor emplacement assembly that allows a robot to autonomously push the TDR waveguides in soil, make a measurement, and retract the waveguides. Additionally, we implement and evaluate multiple adaptive sampling algorithms to build a spatial map of moisture distribution on the field. We use a Gaussian Process (GP) based model to estimate the moisture distribution and report results from simulated and field experiments.

\section{Related works}

Several soil moisture sensing robotic systems have been proposed in recent years, employing different approaches to soil moisture sensing. Broadly, three distinct methods of soil moisture sensing are utilized by autonomous systems: soil sample collection, passive sensing, and in-situ sampling. \textbf{Soil sample collection methods} involve collecting soil samples using a core drill \cite{lukowska2019soil, valjaots2018soil, edulji2023mobile}, scooping  \cite{olmedo2020ugv,anderson2012collecting} etc., which are then analyzed in a controlled environment after extraction, most commonly using the gravimetric (oven drying) method. While this method can provide highly accurate measurements, it results in destroying the sample and is classified as a destructive method \cite{robinson2003review}. It disrupts the natural environment and requires laboratory analysis, making it less suitable for real-time large-scale applications. Passive sensing and in-situ sampling are classified as non-destructive methods. \textbf{Passive sensing} methods are most common in robotic sensing and estimate soil moisture without direct contact with the soil, using techniques such as ground-penetrating radar, hyperspectral imaging, LiDAR, optical camera and other sensors that do not actively manipulate the environment. For instance, Senthil et al. \cite{senthil2018automated} employ a camera sensor and computer vision algorithms to estimate the moisture content of the topsoil. Additionally, Pulido et al. \cite{pulido2020kriging} uses cosmic-ray neutron sensors to measure soil moisture over large areas quickly. However, this approach is both costly and energy-intensive, which limits its practicality for many field applications. The third method, \textbf{in-situ sampling}, involves taking direct measurements at specified locations within the field. These methods typically utilize resistance-based moisture sensors  \cite{ruiz2016ugv}, or capacitance-based moisture sensors \cite{Piper2015, finegan2019development}, which are inserted into the soil robotic manipulators or actuators \cite{linford2024ground, iqbal2020development}. TDR sensors are widely used for soil moisture measurements due to their accuracy in a wide variety of environments and relative ease of usability. Other methods as discussed above, have disadvantages such as lack of accuracy (Resistance Probes \cite{rudnick2015performance}), the need for frequent calibration in different environments (FDR sensors \cite{qin2021analysis}), or a prohibitively large operational cost (Neutron Probe \cite{Kodikara2014} and COSMOS \cite{Kim2008}). TDR sensors offer an efficient trade-off between cost and accuracy. They are relatively resistant to different soil compositions and maintain under 3\% error when calibrated correctly \cite{Bittelli2008}.

Our platform, MoistureMapper, employs in-situ sampling with a TDR sensor mounted on a twin-linear actuator assembly on a ground mobile robot for autonomous sampling and moisture mapping. The work by Dakshinamurthy et al. \cite{dakshinamurthy2024design} relates most closely to this work. They designed a drone equipped with TDR sensors to measure soil moisture. Drones have a severely restricted operational range resulting in limited sensing fidelity. Ground robots offer a larger range of operation, high accuracy and ease of deployment suited for the complex and challenging environment in farms where high canopies may not permit a drone to descend. To the best of our knowledge, there does not exist any prior work in the literature that comprises TDR sensors mounted on ground robots for autonomous soil moisture sensing and mapping.

\subsection*{Adaptive Sampling}
Another contribution of our work is the design, implementation and field deployment of adaptive sampling algorithms for spatial mapping of volumetric moisture in the field using a GP based approach. Adaptive sampling enables the MoistureMapper to efficiently take in-situ soil moisture measurements while minimizing resource consumption. Unlike static sampling strategies, which rely on predefined trajectories or uniform data collection, adaptive sampling dynamically adjusts the sampling process based on current knowledge of the system. This capability is particularly valuable in applications where resources such as energy or time are limited, and where the environment is dynamic or partially unknown. There are two overarching considerations in adaptive sampling: exploration and exploitation. \textit{Exploration} focuses on sampling previously unexplored regions of the environment to develop a broad understanding of the underlying spatial distribution. \textit{Exploitation}, on the other hand, directs sampling toward subregions identified as high-priority through some internal criteria, such as prioritizing areas with high uncertainty or unusual qualities \cite{Rahimi2004}\cite{Andrade-Pacheco2020}. While some adaptive sampling methods rely solely on one of these strategies, most techniques dynamically balance exploration and exploitation in order to gain a comprehensive understanding of the environment. Probabilistic models such as GPs \cite{Westermann2023} and Bayesian \cite{Singh2009} are able to quantify uncertainty and guide sampling decisions toward areas where additional samples would be most beneficial. Recently, adaptive sampling strategies have been employed to optimize kriging models, that relate most closely to our application, by iteratively selecting optimal points based on extracted information \cite{Fuhg2021}\cite{Song2022}. We use a GP based modeling approach and develop acquisition functions for specific optimization criteria. We evaluate the adaptive sampling in simulation and show proof-of-concept with field deployment.

\textit{Paper Organization:} In Section \ref{sec:robot_design} we present MoistureMapper's design and hardware details. In Section \ref{sec:sampling}, we present the spatial mapping problem and solution approach. Section \ref{sec:sim} presents the simulation results and Section \ref{sec:field} gives details of the field deployment. In Section \ref{sec:conclusion}, we discuss future work on the problem. \\\\
\section{MoistureMapper: Robot Design and Hardware}\label{sec:robot_design}

\begin{wrapfigure}{r}{0.5\columnwidth}
    \vspace{-2mm}
    \centering
    \includegraphics[width=0.49\columnwidth]{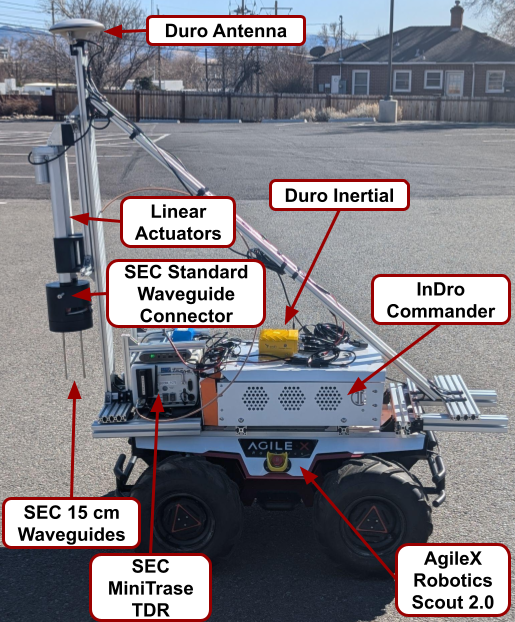}
    \caption{MoistureMapper Hardware Architecture}
    \vspace{-3mm}
    \label{fig:Hardware_Arch}
\end{wrapfigure}

\begin{wrapfigure}{r}{0.5\columnwidth}
    \vspace{-2mm}
    \centering
    \includegraphics[width=0.49\columnwidth]{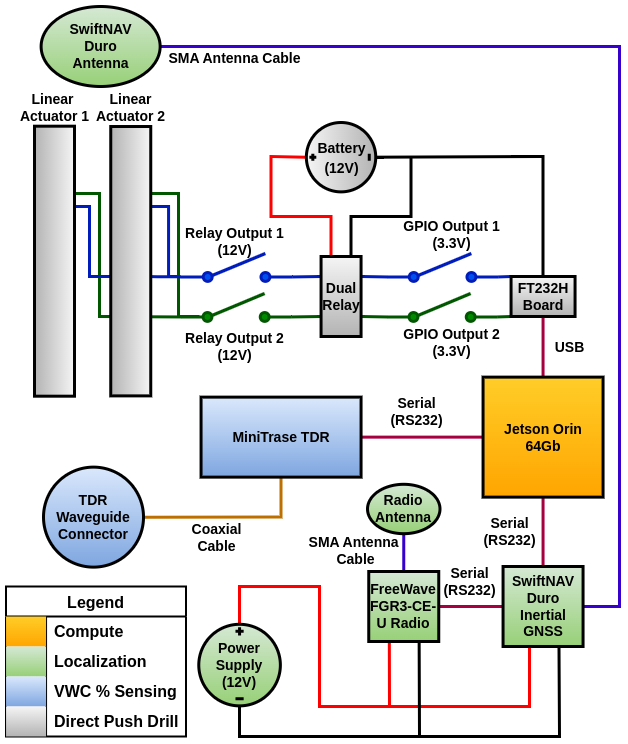}
    \caption{Connection diagram of the MoistureMapper showing electrical components and wiring connections.}
    \vspace{-2mm}
    \label{fig:Wiring_diagram}
\end{wrapfigure}

MoistureMapper is designed to operate in agricultural settings that necessitates a robust and durable design to ensure an extended service life. It is built on the AgileX Robotics Scout2.0 unmanned ground vehicle (UGV) platform, suited for off-road performance. A hardware architecture design for the MoistureMapper is shown in Fig. \ref{fig:Hardware_Arch}. The system is equipped with an onboard Jetson AGX Orin with 64GB RAM.
The Scout UGV is powered by a 30Ah lithium-ion battery, capable of supporting motor operation for approximately 3.5 hours. It is equipped with a PID waypoint navigation controller as part of the onboard autonomy stack. Localization of the MoistureMapper is achieved using a RTK GPS system (SwiftNAV Duro Inertial). A schematic of the robot’s electrical connections is shown in Fig. \ref{fig:Wiring_diagram}.

A Duro Inertial module is mounted on the MoistureMapper system platform, while another Duro module serves as the ground station, enabling RTK GPS localization with accuracy of 0.010 meters. Communication between the Duro modules is facilitated by FreeWave FGR3-CE-U serial radios (902–928 MHz frequency band) paired with Wilson Electronics omnidirectional antennas with a 6.0 dBi gain. This configuration supports a line-of-sight range of up to 60 miles, with the potential for further extension using radio repeaters. Both GNSS modules and radios are powered by an onboard LiPo power source (50000mAh) to ensure sustained operations.

The MoistureMapper system incorporates a mobile TDR sensor, the MiniTrase, developed by SoilMoisture Equipment Corporation (SEC). This sensor offers accurate soil moisture measurements at depths up to 100 cm, with a measurement range of 0–100\% volumetric water content (VWC), a resolution of 10 picoseconds, and an accuracy of $\pm$2\% over the full scale. 
To integrate the MiniTrase with the Jetson AGX Orin an in-house Python library was developed utilizing the sensor's serial protocol to enable autonomous soil moisture data collection. The MoistureMapper system utilizes SEC's standard waveguide connector that supports probe lengths ranging from 8 cm to 20 cm. These components are designed for outdoor deployment. Soil moisture measurements are obtained by driving the TDR sensor's waveguide probes vertically into the soil using two synchronized DC linear actuators. This direct push drill (DPD) mechanism is rigidly mounted onto the Scout platform using 2020 mm aluminum extrusions. A custom 3D-printed mount securely fastens the SEC standard waveguide connector to the actuators. Each linear actuator has a 300 mm stroke length and is capable of exerting up to 3000 N of force, operating in tandem to push the probes into the soil. The DPD system is powered by a dedicated 12V NiCad battery, with voltage regulation controlled by a 12V, 2-channel relay switch module triggered by 3.3V DC outputs from a FT232H module connected to the onboard Jetson AGX Orin.

\section{Moisture Mapping and Adaptive Sensing}\label{sec:sampling}

Given a field with unknown soil moisture levels, our objective is to learn an accurate spatial distribution of volumetric soil moisture in the field using robotic sampling. Each sample measurement incurs the cost of the robot's traversal to the sampling location, as well as the cost of taking the sample measurement (for instance DPD TDR probe deployment cost). We must strategically sample at locations to create an accurate moisture map while minimizing the expended cost. We propose to develop an adaptive sampling method that dynamically adjusts the sampling process based on current knowledge of the system.

\subsubsection*{Problem Statement}
Let the field be a bounded two-dimensional region $X \subset \mathbb{R}^2$. Soil moisture is modeled as an unknown function $f : X \to \mathbb{R}$. At any location \(x \in X\), the robot measures the moisture value using the TDR sensor, $y = f(x)$. However, each sampling sequence comprising insertion of the TDR probe and travel between locations incurs costs. We denote the total cost for sampling at locations $\{x_1, x_2, \ldots, x_M\}$ by $C(D_M) = \sum_{i=1}^{M} c_{\text{sample}}(x_i) + \sum_{i=1}^{M-1} c_{\text{travel}}(x_i, x_{i+1})$. Our objective is to select sample locations that minimize the total cost while building a spatial model \(\hat{f}\) that accurately approximates \(f\) over \(X\).
\subsubsection*{Solution Method}
We develop a Gaussian Process based adaptive sampling strategy. Our sampling strategy proceeds in two phases. In the initial phase, the robot collects measurements on a coarse grid to develop a preliminary model. In the subsequent iterative phase, the sampling process is guided by a Gaussian Process (GP) that provides predictive mean and uncertainty estimates.

\subsubsection*{Initial Coarse Sampling} 

The field \(X\) is partitioned into a coarse grid. The robot collects initial measurements at these grid points: $ D_0 = \{(x_i, f(x_i)) \mid i=1,\ldots,M_0\}$. A GP is trained on \(D_0\) using a zero mean and the Squared Exponential (RBF) kernel: $k(x,x') = \sigma^2 \exp\!\left(-\frac{\|x - x'\|^2}{2\ell^2}\right)$, where \(\sigma^2\) is the signal variance (set to 1 for normalization) and \(\ell\) is the length scale. The GP yields, for each \(x \in X\), a predictive mean \(\mu(x)\) and a predictive variance \(\sigma^2(x)\), with the latter serving as a measure of uncertainty. The predictive mean and variance are computed as \( \mu(x) = k(x, X) [K(X, X) + \sigma_n^2 I]^{-1} \mathbf{y} \) and \( \sigma^2(x) = k(x, x) - k(x, X) [K(X, X) + \sigma_n^2 I]^{-1} k(X, x) \), respectively. \(k(x, X)\) is the vector of covariances between $x$ and \(X = [x_1, \ldots, x_{M_0}]\), \(K(X, X)\) is the covariance matrix, \(\sigma_n^2\) is the noise variance, \(I\) is the identity matrix, and \(\mathbf{y} = [f(x_1), \ldots, f(x_{M_0})]^\top\) is the set of observed moisture values.

\begin{figure*}[htbp]
    \centering
    \hspace*{-0.9cm}
    \subfloat[uniform field]{\includegraphics[width=0.3\textwidth]{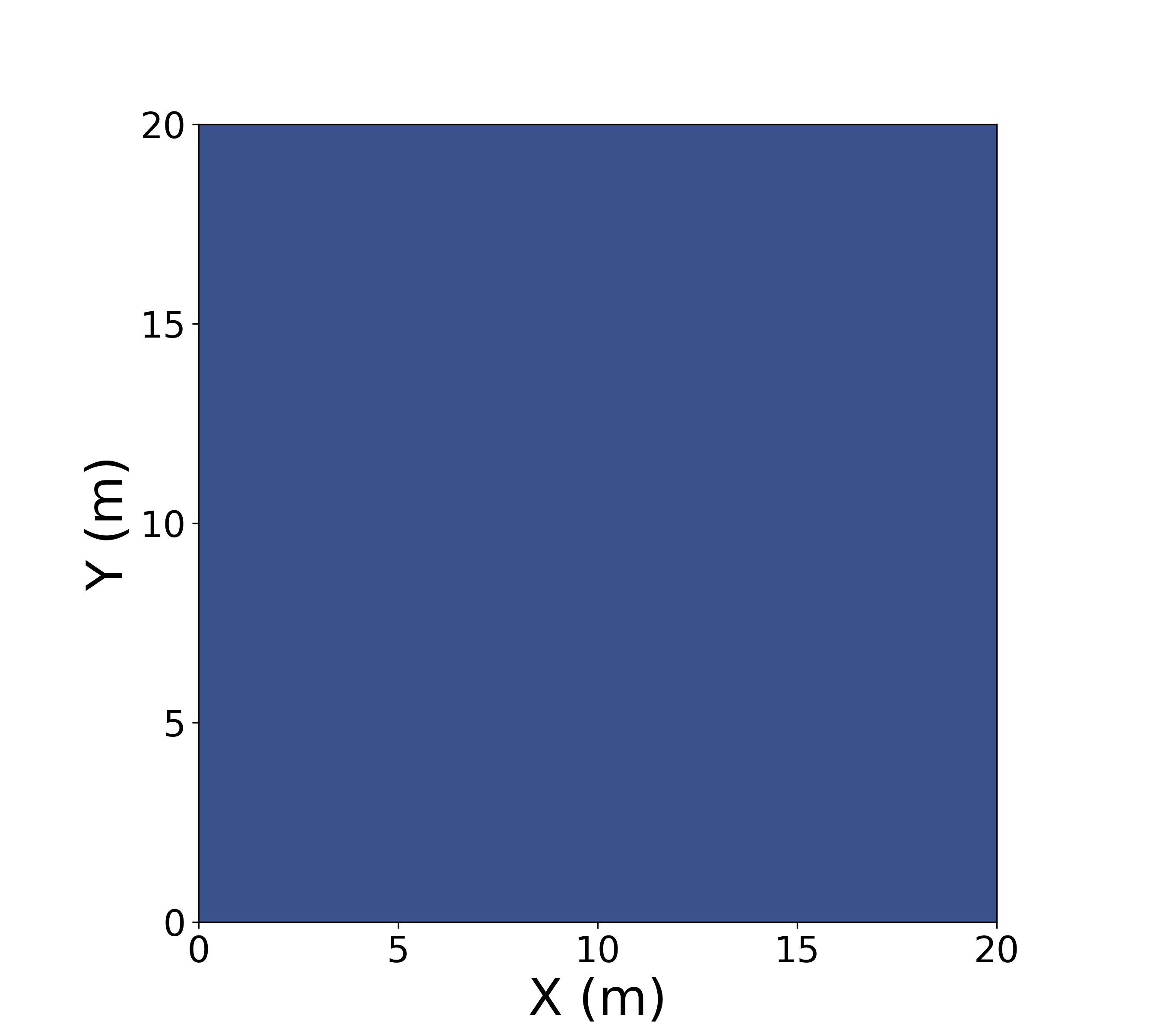}}
    \hspace*{-0.8cm}
    \subfloat[sloped field]{\includegraphics[width=0.233\textwidth]{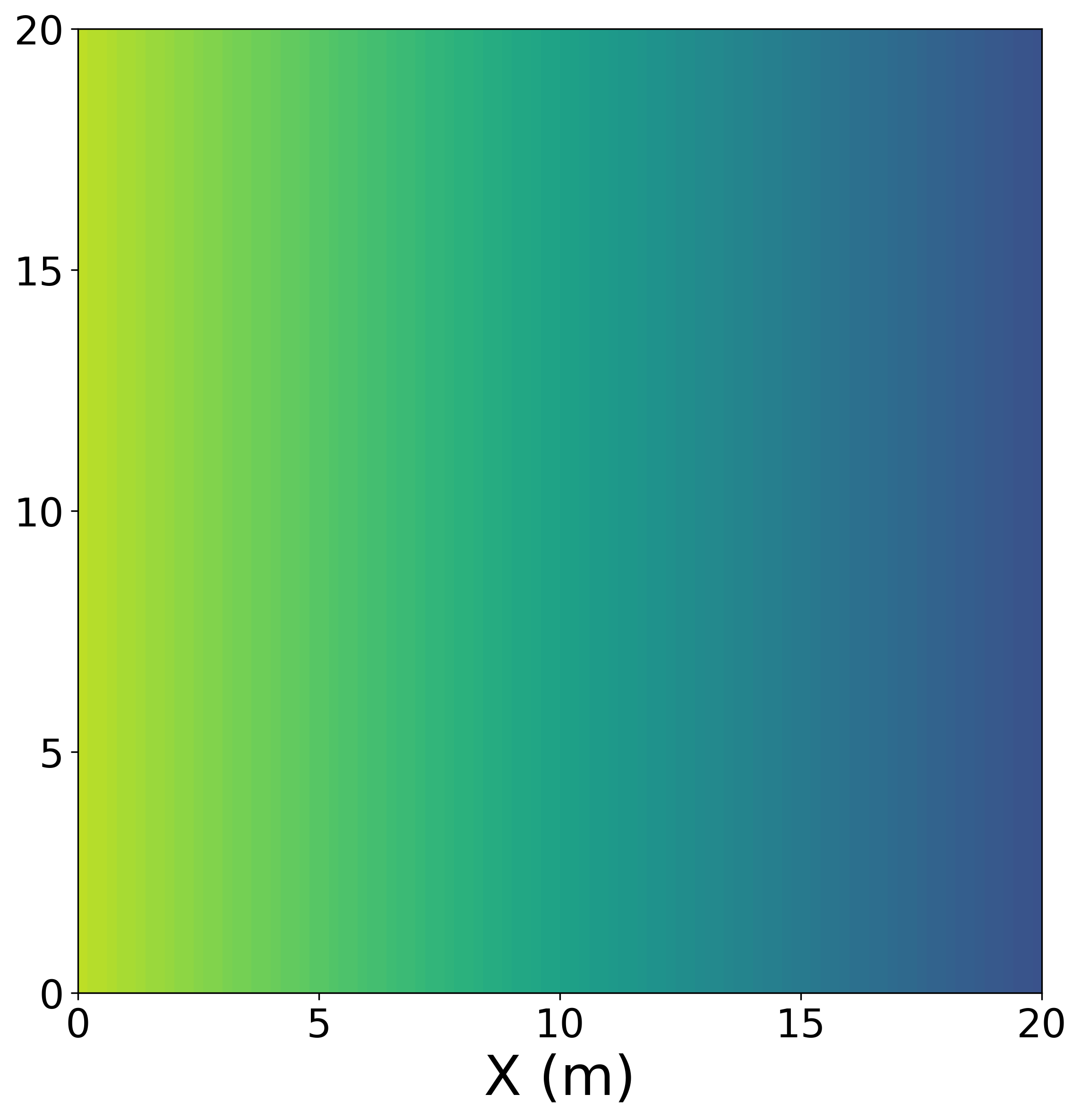}}
    \hspace*{-0.15cm}
    \subfloat[Gaussian field]{\includegraphics[width=0.233\textwidth]{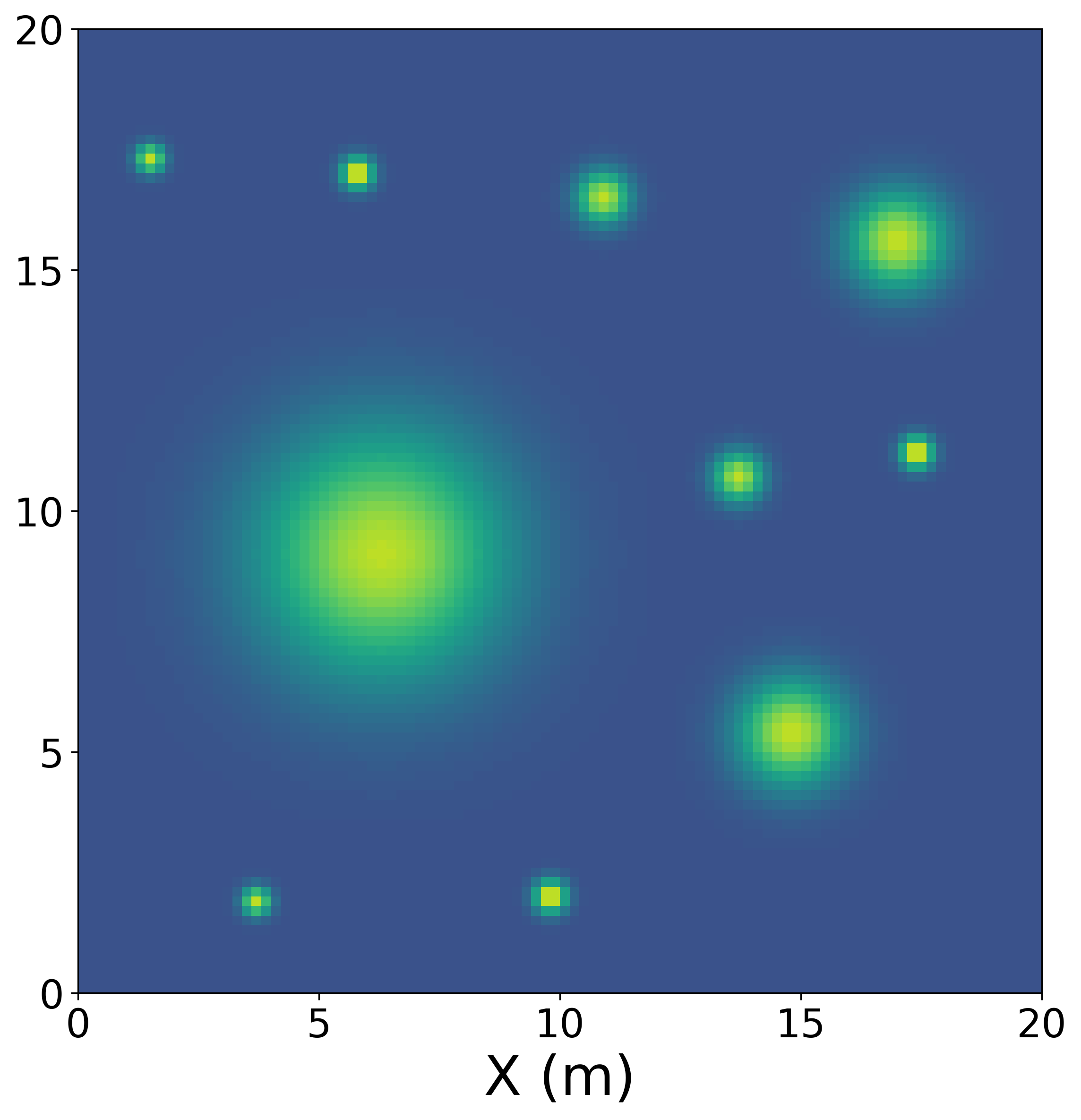}}
    \hspace*{-0.15cm}
    \subfloat[hybrid (sloped and Gaussian)]{\includegraphics[width=0.267\textwidth]{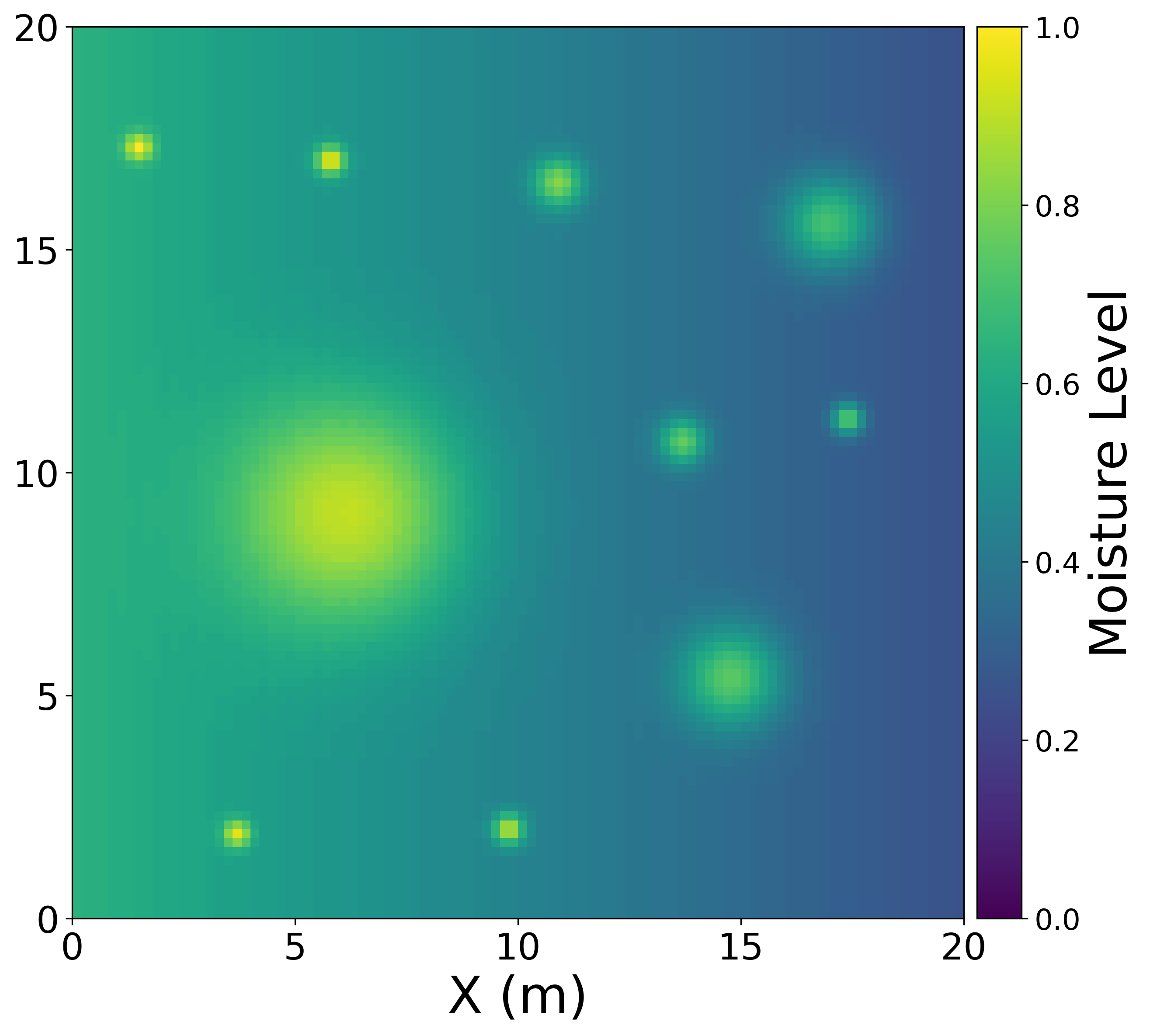}}
    \caption{Sample instances of the four different moisture distribution maps used in simulations.}
    \label{fig:field_types}
\end{figure*}
\subsubsection*{Iterative Adaptive Sampling} Once the initial GP is trained, the robot iteratively refines its moisture map by selecting new sampling locations that are both informative and cost-effective. In iteration \(m\), the current GP (trained on dataset \(D_m\)) predicts for any candidate point \(x\) the moisture level \(\mu_m(x)\) and the uncertainty \(\sigma_m^2(x)\). To simultaneously account for the benefit of sampling in regions of high uncertainty and the cost of traveling long distances, we propose two sample acquisition functions: 
\begin{itemize}
    \item $A_1(x) = \sigma_m^2(x) \left( 1 - \frac{d(x_{\text{current}}, x)}{d_{\text{max}}} \right)$
    \item $A_2(x) = 0.5*\left(\sigma_m^2(x) + \left( 1 - \frac{d(x_{\text{current}}, x)}{d_{\text{max}}} \right)\right)$, where,
\begin{itemize}
    \item \(\sigma_m^2(x)\) is the predictive variance at \(x\),
    \item \(d(x_{\text{current}}, x)\) is the Euclidean distance from the robot's current location \(x_{\text{current}}\) to \(x\),
    \item \(d_{\text{max}}\) is the maximum distance between two points in the environment.
\end{itemize}
\end{itemize}

These proposed acquisition functions, \( A_1(x) \) and \( A_2(x) \), provide two distinct strategies for iterative adaptive sampling in robotic moisture mapping. The first function, \( A_1(x) \),
employs a multiplicative structure to balance uncertainty \( \sigma_m^2(x) \) and travel cost, favoring sampling locations that simultaneously exhibit high uncertainty and proximity to the current position of the robot. In contrast, \( A_2(x) \) utilizes an additive structure, treating uncertainty and cost efficiency as independent objectives. This allows for greater flexibility, as high-uncertainty regions can compensate for increased travel distances, promoting broader exploration of the environment.

For both acquisition functions, the next sample location is chosen by maximizing the acquisition function value over the domain $X$, $ x_{m+1} = \arg\max_{x \in X} A_i(x)$. The robot travels from its current location \(x_m\) to \(x_{m+1}\). At \(x_{m+1}\), the robot collects a moisture measurement \(f(x_{m+1})\) using the onboard TDR sensor. At this point, the new measurement is added to the dataset, $D_{m+1} = D_m \cup \{(x_{m+1}, f(x_{m+1}))\}$ and the GP is updated with \(D_{m+1}\) to refine its predictions. In order to prevent \( A_1(x) \) and \( A_2(x) \) from getting stuck in local minima, we add a randomized component, wherein the next sample location is chosen randomly from amongst the top five candidate locations. The iterative process continues until a stopping condition is met, either the maximum number of iterative samples, $\eta$ have been collected, the maximum distance traveled while collecting samples exceeds the threshold, $\delta$, or the overall predictive variance across \(X\) falls below a predefined threshold, $\psi$. 

As a baseline, we also implement a benchmark sampling algorithm that selects the next sample location as the location with the highest predictive variance. This greedy strategy maximizes information gain without consideration for travel cost, allowing for a useful comparison against the proposed acquisition functions.
\section{Simulation Results}\label{sec:sim}

We developed a custom simulator for the sampling algorithm in Python. We use a square environment of side `\textit{s}' and varied the environment size from s = 20 to 100 in steps of 20. For each environment size we generated 12 unique maps : 1 uniform, 1 gradient, 5 Gaussian, and 5 hybrid maps, resulting in a total of 60 maps. The hybrid map comprises a combination of gradient and Gaussian clusters. Sample environments for each distribution are shown in Fig. \ref{fig:field_types}. For the Gaussian and hybrid maps, 10 clusters were generated per scenario with each cluster’s radius randomly chosen from the interval $[1, s/10]$. The simulations were executed on a Dell Latitude 5420 laptop with i5 (2.60GHz × 8) processor with 16 GB RAM. Three stopping criteria were used to compare the adaptive sampling algorithms presented in Section \ref{sec:sampling}:
\begin{itemize}
    \item \textit{\# Sample:} [20,40,60,80,100]
    \item \textit{Distance Traveled:} [300,600,900,1200,1500]
    \item \textit{Maximum Uncertainty:} {0.4}
\end{itemize}

\begin{figure*}[htbp]
    \centering
    \subfloat{\label{fig:avg_dist_vs_env}\includegraphics[width=0.24\textwidth,trim=25 10 45 20,clip]{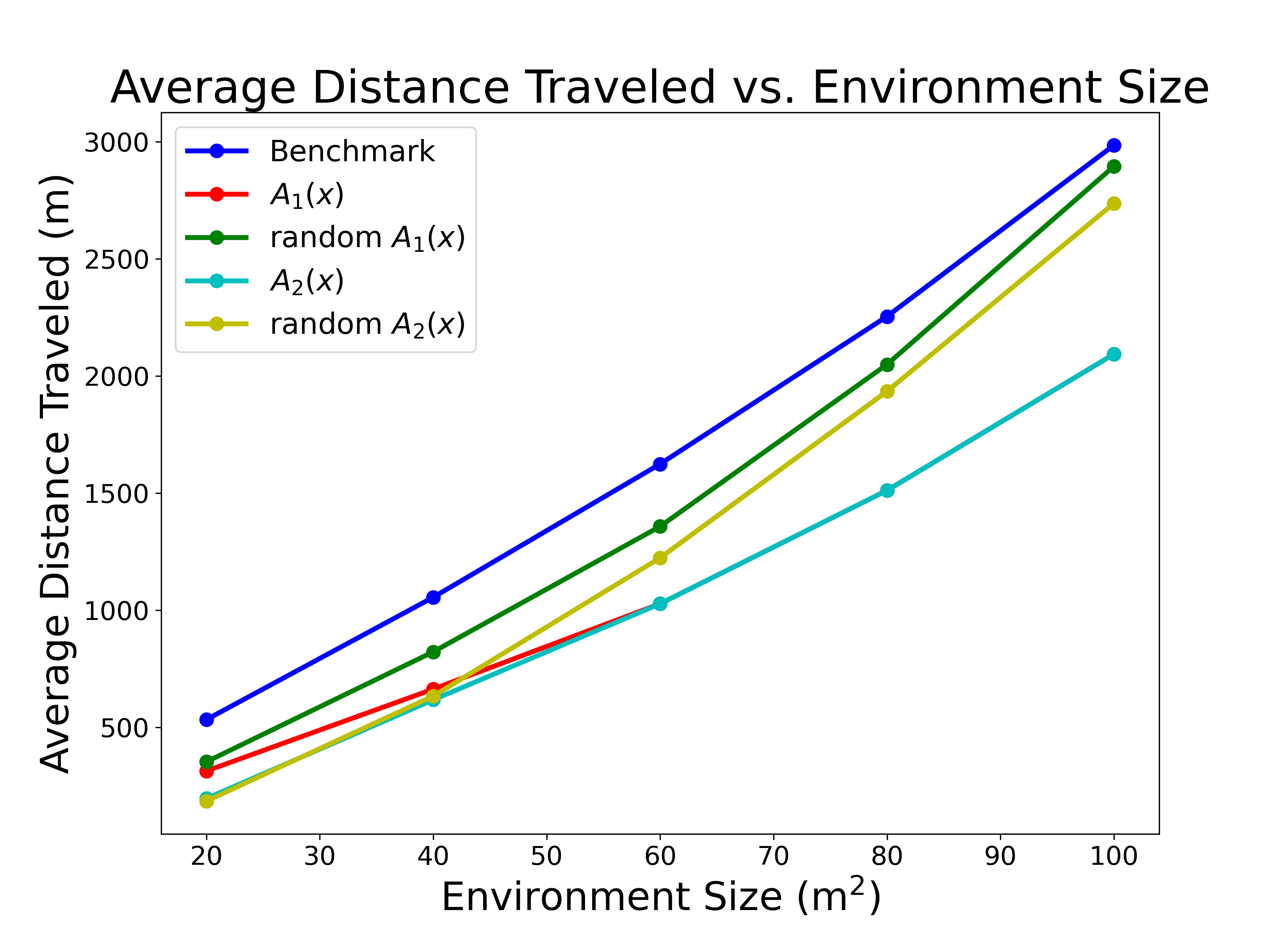}}  \subfloat{\label{fig:avg_sample_vs_env}\includegraphics[width=0.23\textwidth,trim=25 10 45 20,clip,trim=30 10 70 20,clip]{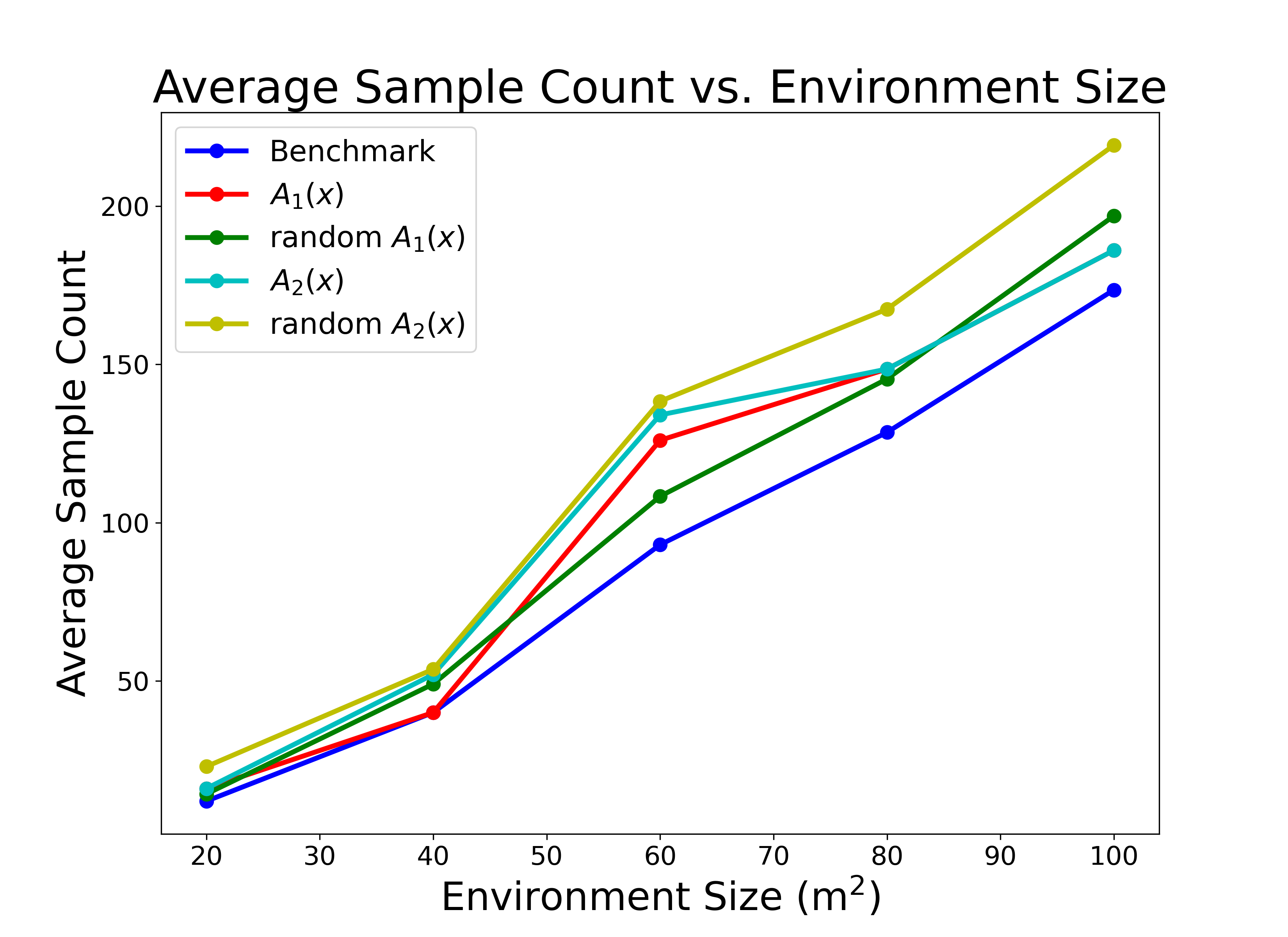}}    \subfloat{\label{fig:avg_max_var_vs_env}\includegraphics[width=0.24\textwidth,trim=25 10 45 20,clip,trim=25 10 40 20,clip]{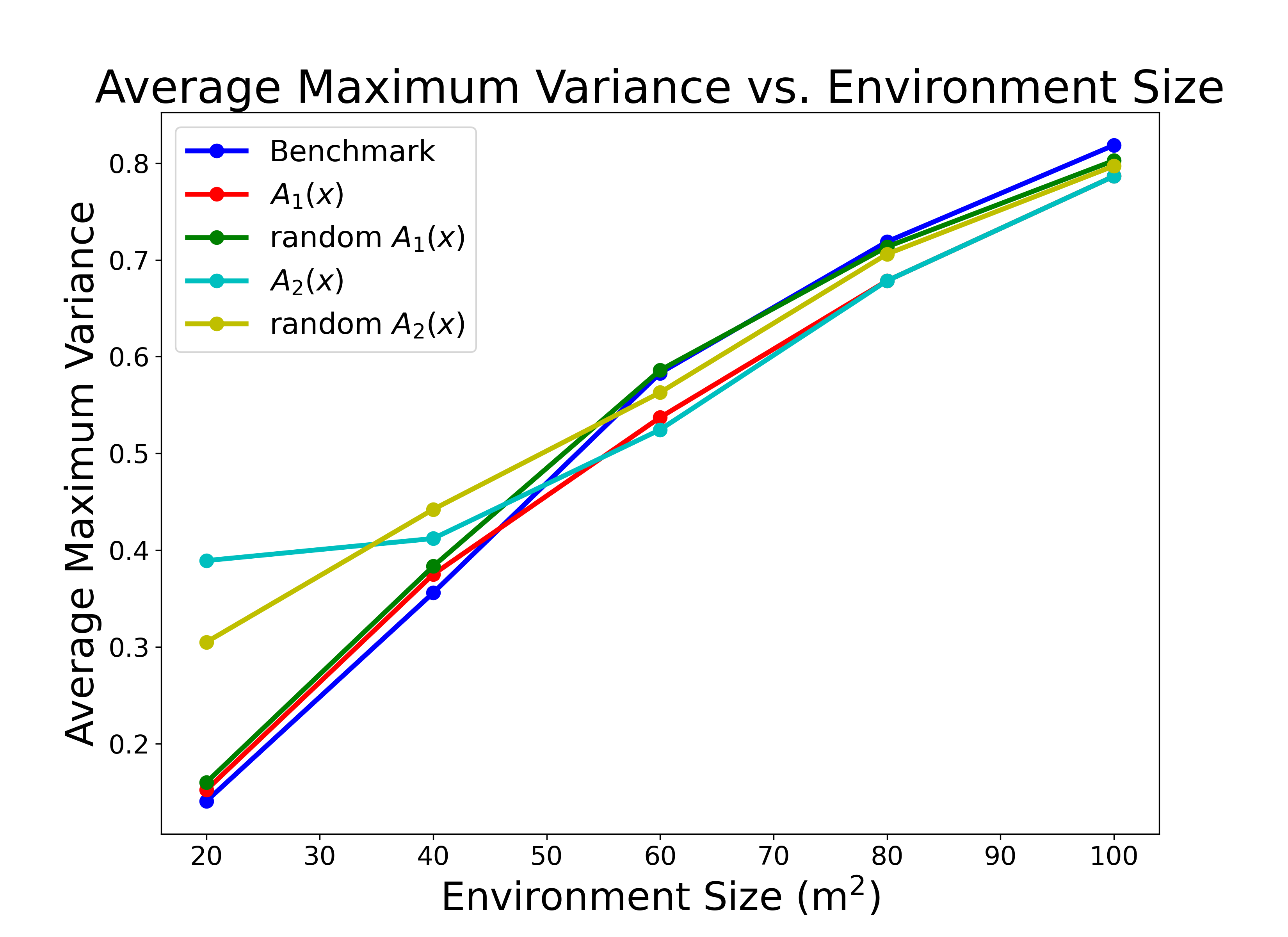}}    \subfloat{\label{fig:avg_avg_var_vs_env}\includegraphics[width=0.24\textwidth,trim=25 10 45 20,clip,trim=25 10 50 20,clip]{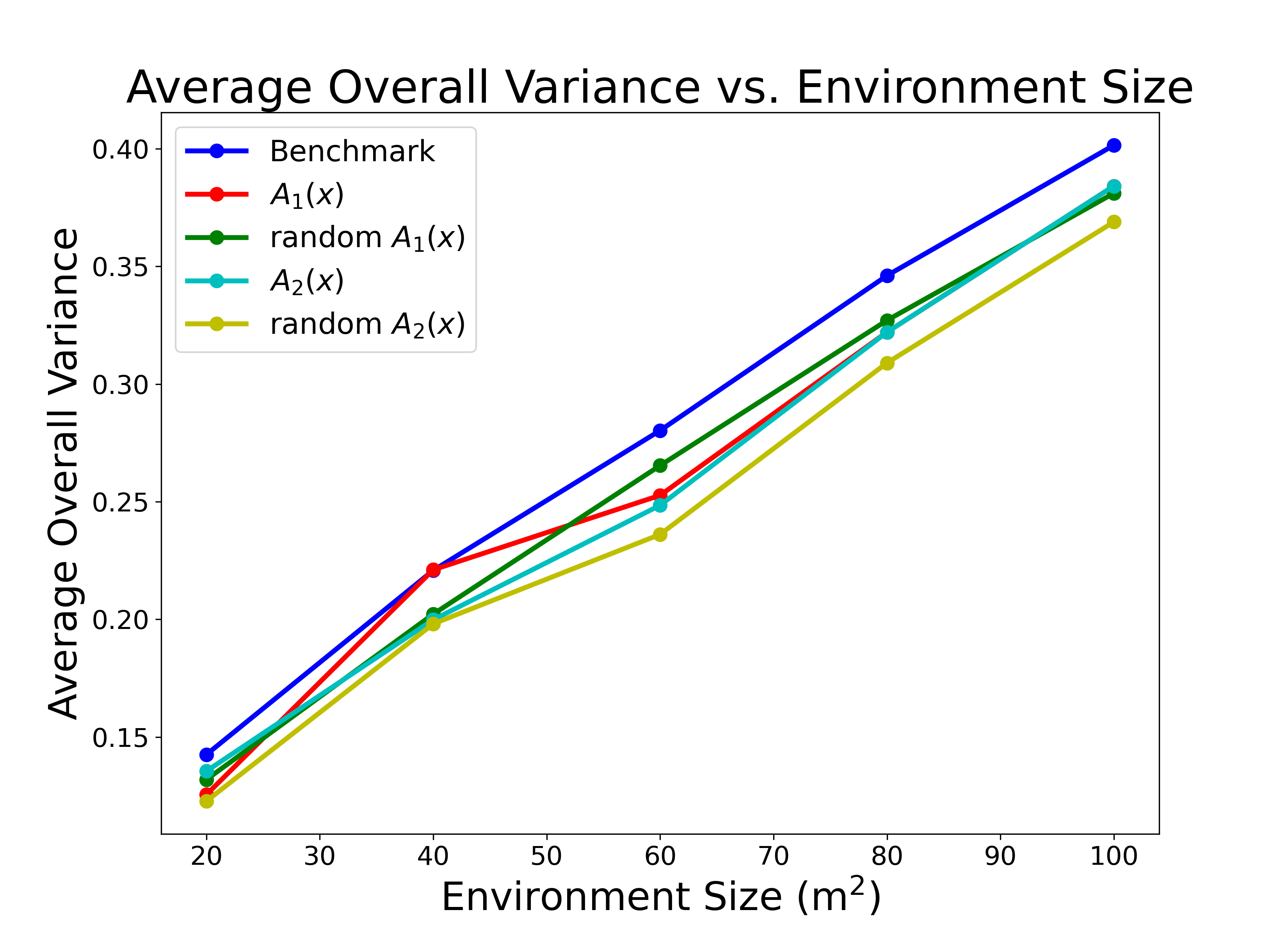}}
    \caption{Average metrics across all simulations, compared across different environment sizes}
    \label{fig:simulation_metrics}
\end{figure*}

\subsection*{Results} 

Fig. \ref{fig:simulation_metrics} presents the average results for each sampling approach across increasing environment sizes. Each simulation scenario includes 12 randomized moisture maps per environment size. Metrics reported include total distance traveled (Fig. \ref{fig:simulation_metrics}a), number of samples (Fig. \ref{fig:simulation_metrics}b), maximum residual variance (Fig. \ref{fig:simulation_metrics}c), and average residual variance (Fig. \ref{fig:simulation_metrics}d). Plots (a), (b) and (c) exclude instances where the stopping criterion is conditioned on the dependent variable.

The benchmark algorithm results in the highest total distance traveled across all field sizes, as shown in Fig. \ref{fig:simulation_metrics}a. This is expected due to its greedy sampling behavior that selects the location with the highest predictive variance without considering travel cost. $A_1$ and $A_2$ reduce travel distance by incorporating distance penalties in their acquisition functions, with $A_2$ showing up to 30\% improvement in travel distance over the benchmark. The randomized variants of both $A_1$ and $A_2$ result in larger values for distance traveled, reflecting occasional longer detours induced by random selection.

Fig. \ref{fig:simulation_metrics}b reports the sample count for each method. The benchmark consistently collects the least number of samples, since high-variance sampling locations result in fastest reduction of maximum system variance. Adaptive methods collect more samples, with randomized $A_2$ showing the highest sample counts due to increased exploration. These trends are amplified in larger environments, where spatial coverage results in larger travel distance for each sample.

Maximum variance values at the end of the sampling plan are shown in Fig. \ref{fig:simulation_metrics}c. While the benchmark achieves low maximum variance in small environments, its performance degrades with increasing field size. $A_1$ and $A_2$ have slightly lower values for maximum variance in larger fields, with randomized variants showing more variability. The plot shows that all strategies show similar behavior in large environments.

Fig. \ref{fig:simulation_metrics}d presents the average variance of the environment at the end of the sampling plan. Adaptive methods consistently outperform the benchmark in reducing overall field uncertainty. Notably, randomized $A_2$ achieves the lowest average variance across all sizes, indicating improved spatial coverage. Even in runs terminated by sample count or distance thresholds, average variance remains below 0.4 in most cases and with up to 5\% improvement over the benchmark, indicating that adaptive planners can achieve better uncertainty reduction.

These results demonstrate the trade-offs between sample efficiency, spatial coverage, and estimation quality. The benchmark offers faster local reduction of maximum variance but suffers from excessive travel and higher average variance values. $A_1$ offers a favorable balance between distance travel and sample count, while $A_2$ and its randomized variant achieve more uniform uncertainty reduction at higher cost. Method selection should depend on deployment constraints and whether minimizing distance traveled, sample count, or overall uncertainty is the primary objective.

\begin{figure}[htbp]
    \centering
    \captionsetup[subfigure]{font=normal}
    \subfloat[Ground truth]{\includegraphics[width=0.24\textwidth]{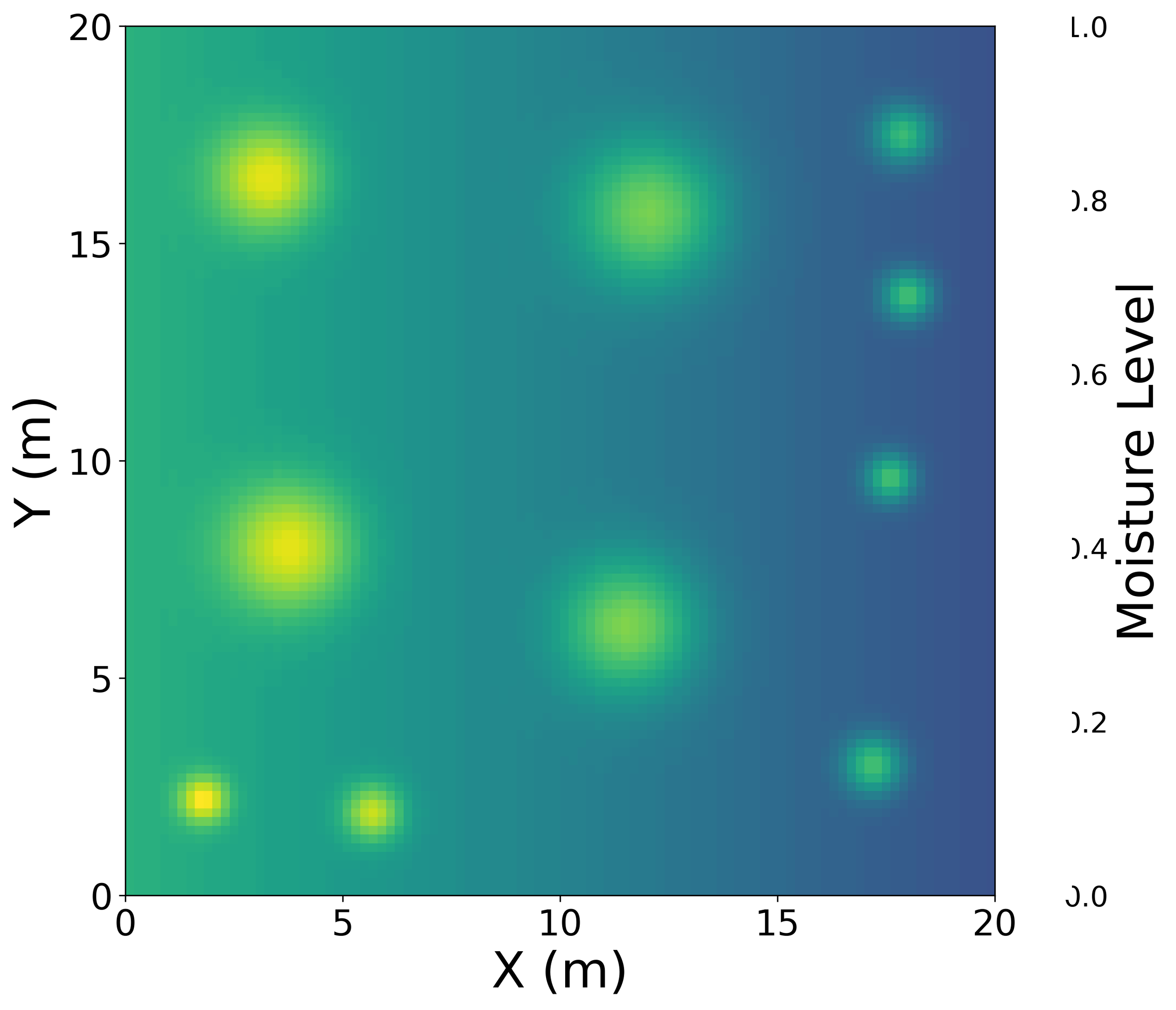}} 
    \subfloat[Reconstruction]{\includegraphics[width=0.23\textwidth]{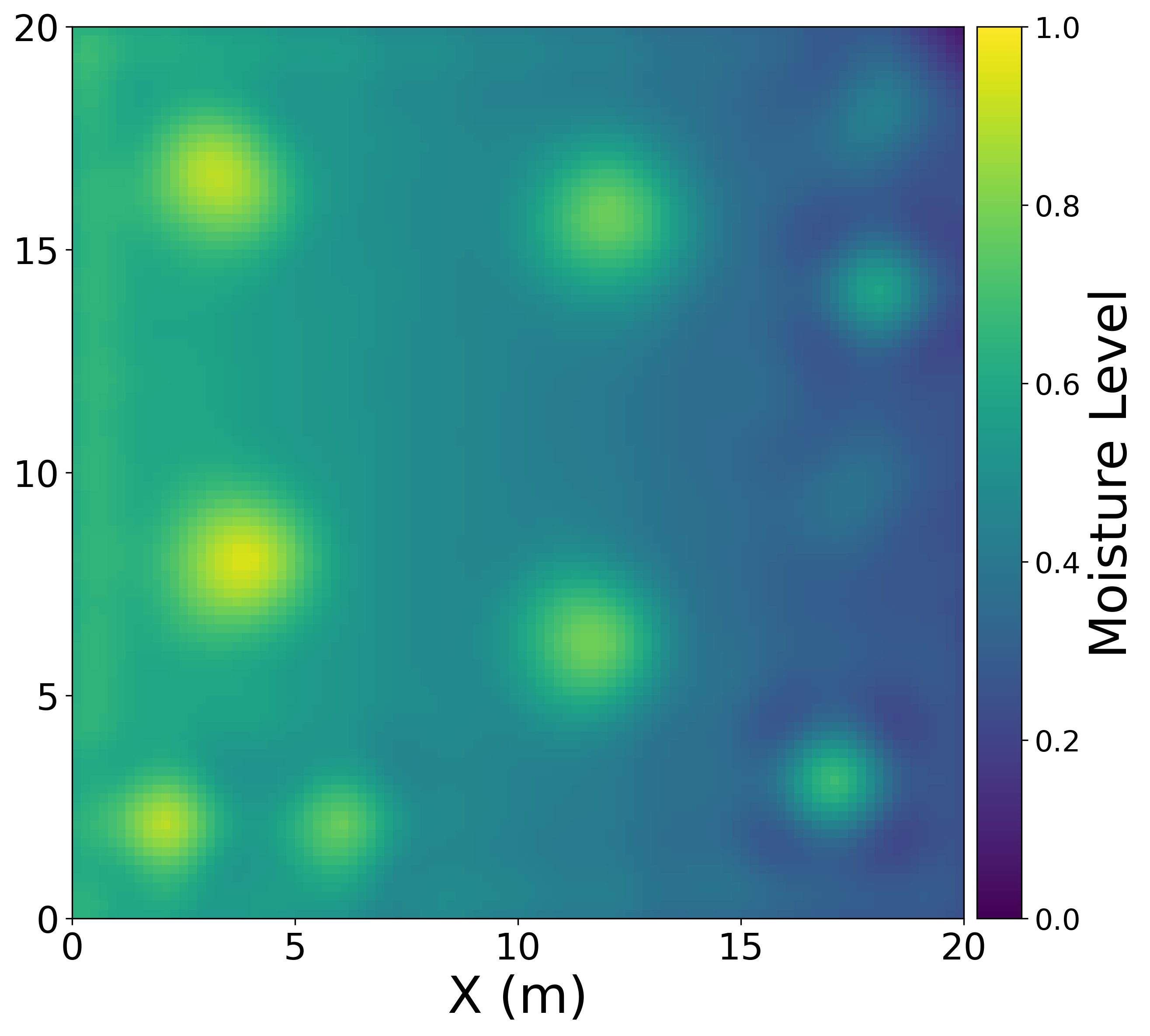}}
    \caption{Comparison of a simulation-reconstructed map (\( A_1(x) \), $\psi=0.4$) to the ground truth generated map}
    \label{fig:reconstruction}
\end{figure}

After each sampled point, a moisture distribution map of the environment was reconstructed using Gaussian Process Regression (GPR), based on the available data points, with the underlying assumption that the moisture distribution within the field follows a spatial Gaussian distribution. Fig. \ref{fig:reconstruction} shows a sample instance.

\section{Experimental Evaluation}\label{sec:field}
In this section, we present results from evaluations of the MoistureMapper in laboratory and field deployments.

\subsection{Laboratory Test}
Before field deployment, the TDR on the MoistureMapper was verified for performance in the lab using soil buckets. The measurements made by MoistureMapper are compared to the ground truth (GT) measurements taken manually by the SEC HandiTrase system. Both systems were equipped with 15cm waveguides. Each set of measurements was preceded by a manually measured amount of water added to a soil bucket. The results of this test are given in Table \ref{tab:Lab_Results} and are shown to have an average VWC error of 0.7\%.

\begin{table} [hbt!]
    \centering
    \caption{Lab Test Readings}
    \begin{tabular}{|c|c|c|c|c|}
        \hline
        \multirow{2}{4em}{Liter(s) Added} & \multicolumn{2}{c|}{Volumetric Water $\%$ } & \multicolumn{2}{c|}{Dielectric Constant}  \\
        \cline{2-5}
        & GT & MoistureMapper & GT & MoistureMapper \\
        \hline
        0.0 & 7.4 & 7.1 & 4.9 & 4.7 \\
        0.5 & 10.6 & 10.0 & 6.2 & 6.0 \\
        1.0 & 25.1 & 25.5 & 12.9 & 13.3 \\
        1.5 & 26.7 & 26.1 & 14.3 & 13.8 \\
        2.0 & 28.8 & 29.6 & 16.3 & 17.1 \\
        2.5 & 29.0 & 30.1 & 16.5 & 17.5 \\
        3.0 & 32.3 & 33.4 & 19.2 & 20.0 \\
        \hline
    \end{tabular}
    \label{tab:Lab_Results}
\end{table}

\subsection{Field Experiments} 

\begin{figure}[htbp]
    \centering
    \subfloat[VWC Measurements]{\includegraphics[width=0.24\textwidth]{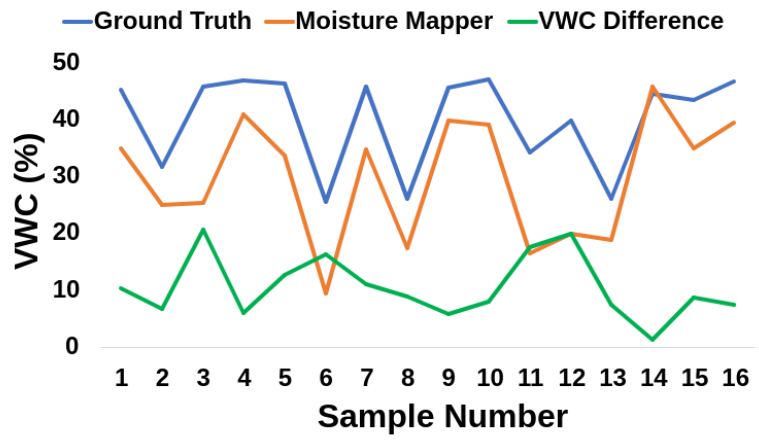}} 
    \hspace*{0.0cm}
    \subfloat[Trajectory]{\includegraphics[width=0.24\textwidth]{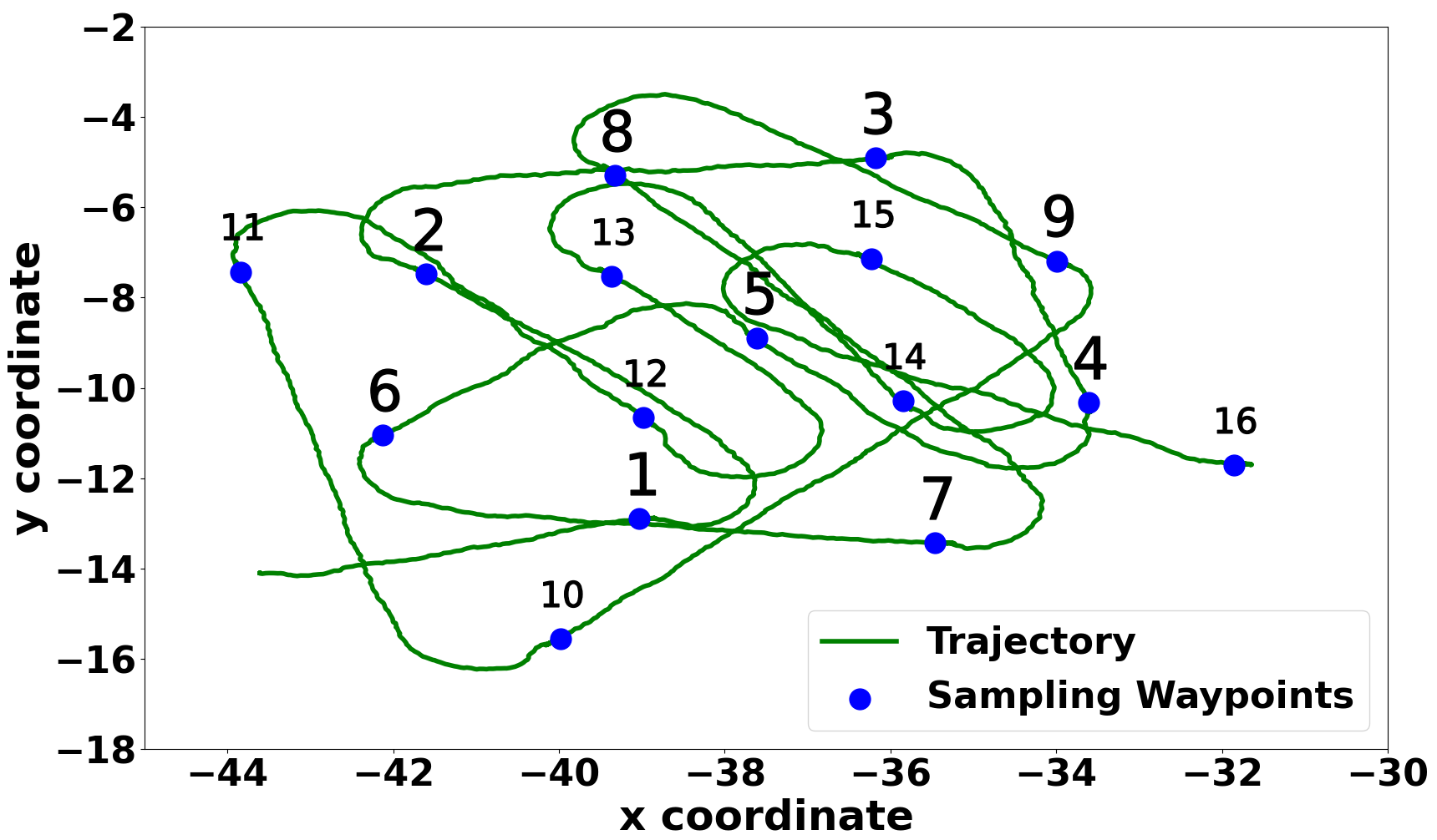}}
    \caption{Field results for the benchmark sampling algorithm.}
    \label{fig:Field_Moisture_Values_benchmark_2}
\end{figure}

\begin{figure}[htbp]
    \centering
    \subfloat[VWC Measurements]{\includegraphics[width=0.24\textwidth]{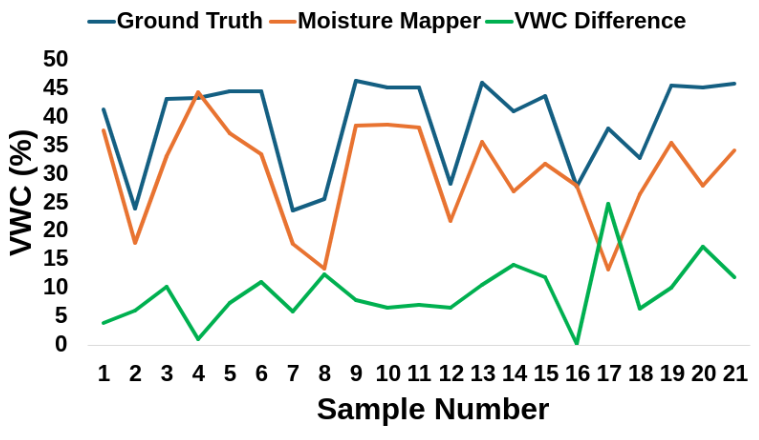}} 
    \hspace*{0.0cm}
    \subfloat[Trajectory]{\includegraphics[width=0.24\textwidth]{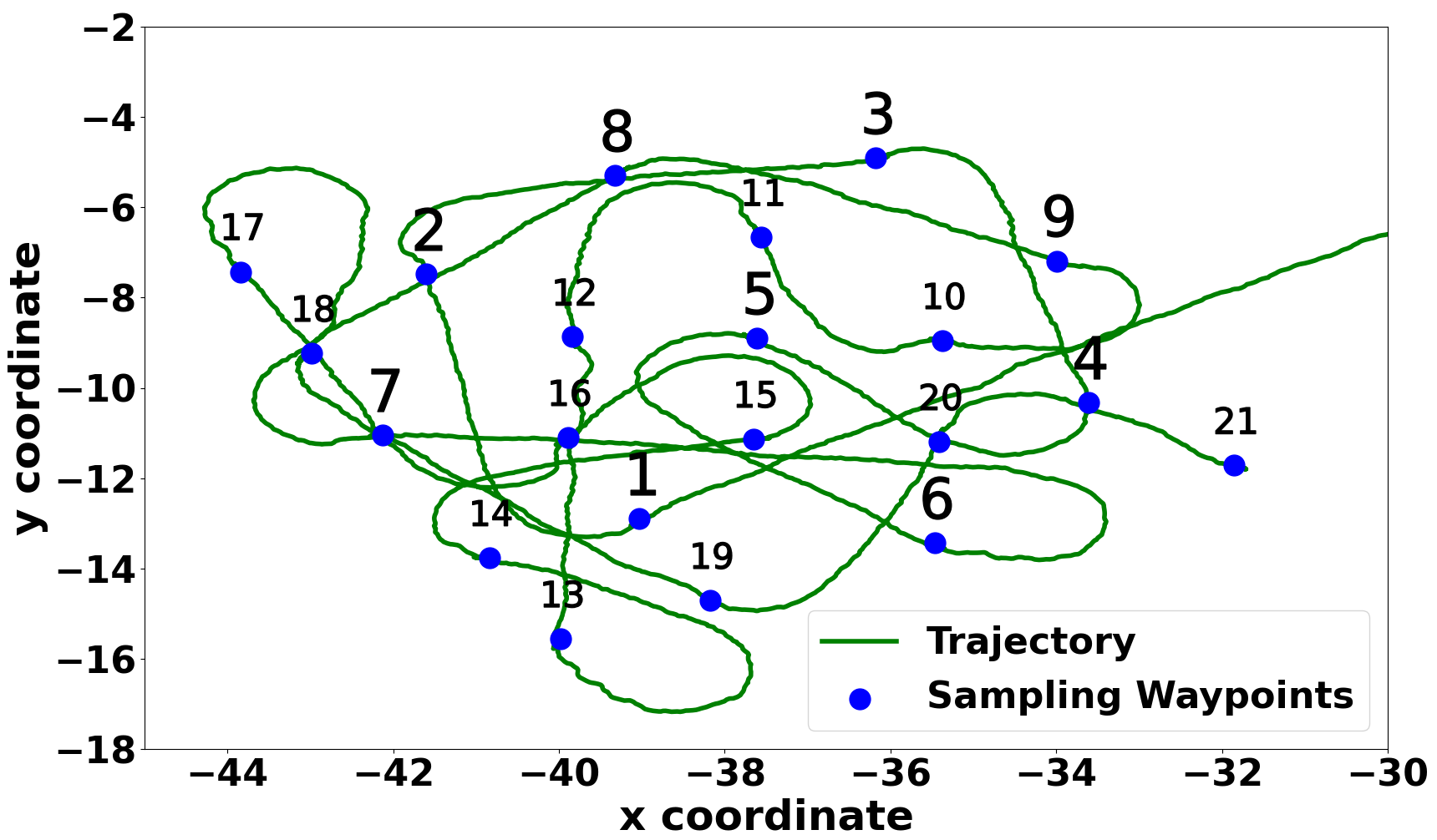}}
    \caption{Field results for the $A_1$ adaptive sampling algorithm.}
    \label{fig:Field_Moisture_Values_a1}
\end{figure}

We conducted a field evaluation of the MoistureMapper at Dick Taylor Memorial Park (Reno, NV), in a 10 × 10 meter grass field. In the field experiment, the benchmark and $A_1$ adaptive sampling algorithms were tested. For each sampling test, Phase 1 of the adaptive planner generated four waypoints within the designated sampling area to compute an initial moisture map of the field. The adaptive planner then recursively determined the next waypoint based on the selected sampling algorithm. The waypoint navigation system guided the robot to the sampling location. Once the MoistureMapper reached the target location, the DPD system was activated to insert the 15 cm waveguides into the soil and make a measurement. Following the measurement, the planner updated the variance grid to determine the next sampling location until the stopping criterion was met (i.e., maximum system entropy below a threshold of 0.2). VWC and dielectric constant were also measured manually using an Acclima TDR-315H (15 cm probes) at each sampling location to record the ground truth (GT). Measured moisture values and robot trajectories for the benchmark and $A_1$ sampling tests are shown in Fig. \ref{fig:Field_Moisture_Values_benchmark_2} and Fig. \ref{fig:Field_Moisture_Values_a1}, respectively.


Table \ref{tab:Field_Sampling_Table} reports observations from the field experiments. The benchmark algorithm required 16 samples, while the $A_1$ adaptive planner required 21 samples to reduce system variance to the same value. Total distance traveled and total mission time were higher for the $A_1$ planner at 176 m and 2607 s, compared to 153 m and 2019 s for the benchmark planner. Since the $A_1$ planner prioritizes sampling locations closer to the robot's current location, the average travel distance per sample was expectedly lower at 8.4 m, compared to 9.5 m for the benchmark.


\begin{table} [hbt!]
    \centering
    \caption{Field Experiment Statistics}
    \begin{tabular}{|c|c|c|}
        \hline
        & Benchmark & A1 \\
        \hline
        \# of Samples & 16 & 21 \\
        Total Distance Moved (m) & 153.30088 & 176.41359 \\
        Distance per sample (m) & 9.58131 & 8.40065 \\
        Total sampling time (s) & 2019.3 & 2607.9 \\
        \hline
    \end{tabular}
    \label{tab:Field_Sampling_Table}
\end{table}

It was observed that the MoistureMapper’s moisture measurements were consistently lower than the ground truth values. This discrepancy is attributed to the fact that the MoistureMapper could not insert the full length of the probes at many sampling locations due to excessively dense soil which caused the front of the UGV to raise into the air instead of driving the probes downward. To account for the inability to reliably achieve full waveguide insertion in all areas, tick marks were made on the waveguides at each centimeter increment. During each moisture measurement sequence, the number of tic marks visible above the soil surface was manually counted and recorded. The depth of waveguide insertion into the soil was then calculated for each sample.  
\begin{wrapfigure}{r}{0.5\columnwidth}
    \centering
    \includegraphics[width=0.49\columnwidth]{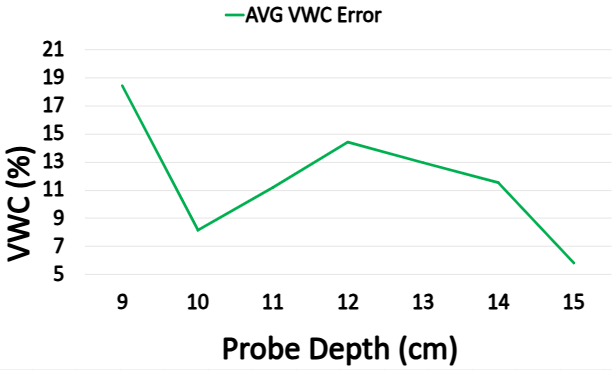}
    \caption{Average error in VWC \% measurements by MoistureMapper relative to the probe depth.}
    \label{fig:probe_depth_errors}
\end{wrapfigure}
Fig. \ref{fig:probe_depth_errors} shows the variation in average VWC error with the length of the probe inserted in the soil. The plot shows improved accuracy with increase in inserted probe length. It was observed that TDR probe accuracy can drastically reduce even with little exposure to air. This is due to the high dielectric of water saturated soil (5-40+) compared to air (1).

\begin{figure}[htbp]
    \centering
    \subfloat[Ground Truth]{\includegraphics[trim=20 0 170 65, clip, width=0.155\textwidth]{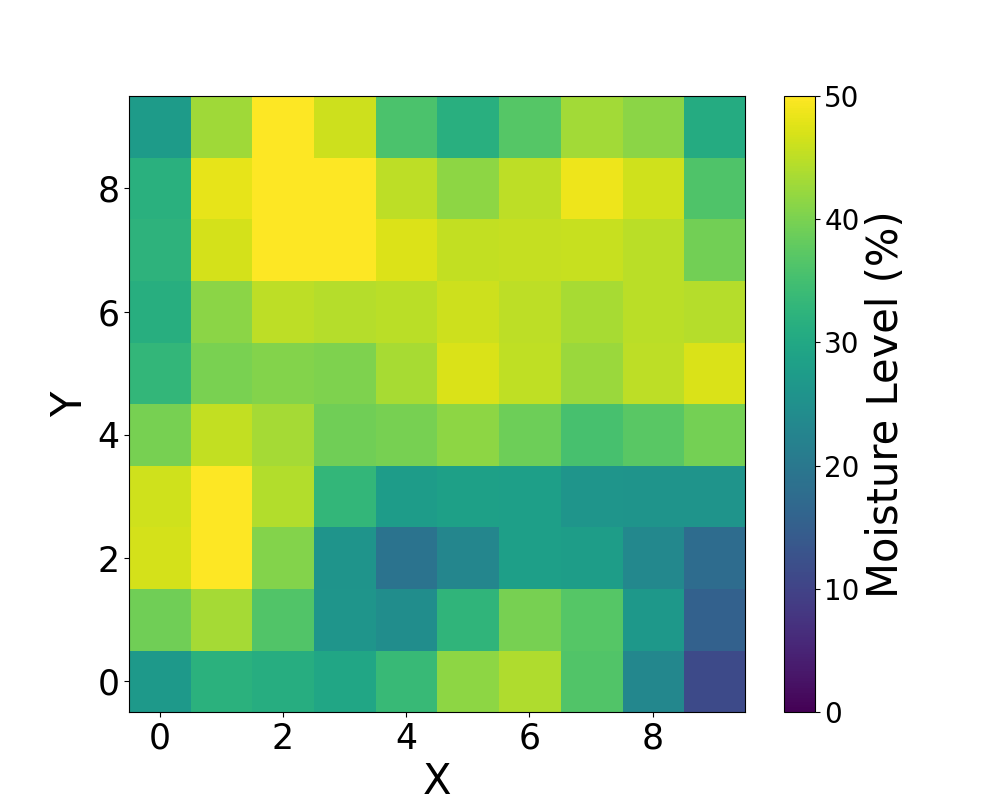}} 
    \subfloat[Benchmark]{\includegraphics[trim=60 0 170 65, clip, width=0.145\textwidth]{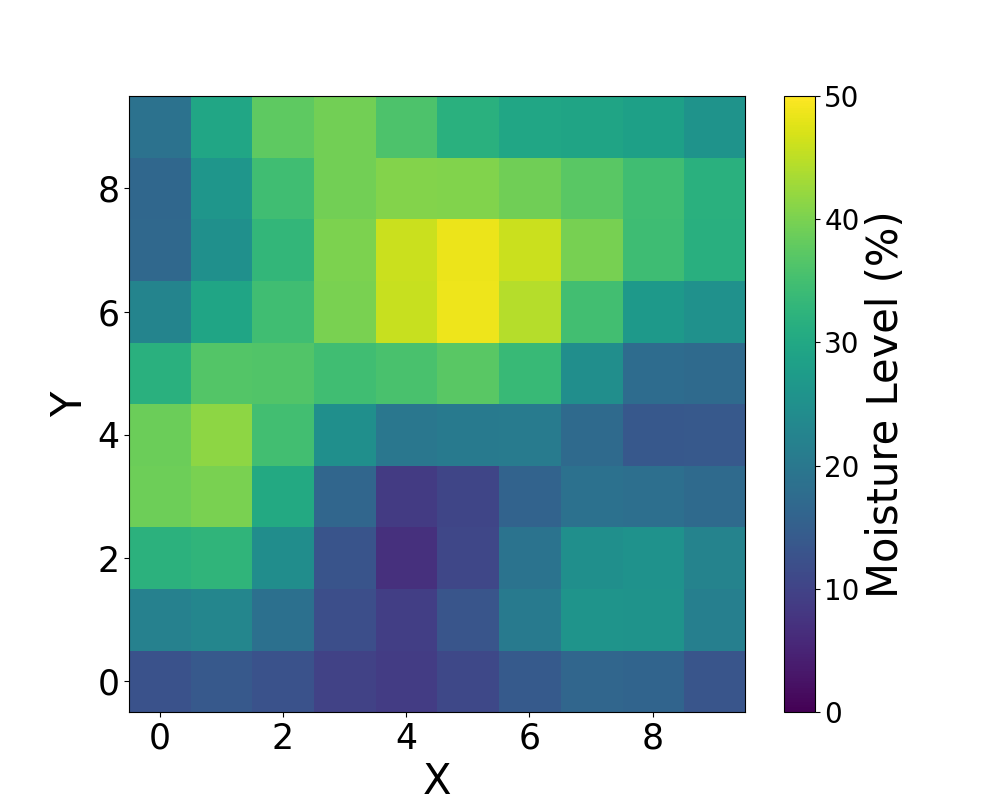}}
    \subfloat[$A_1$ planner]{\includegraphics[trim=60 0 20 64.5, clip, width=0.19\textwidth]{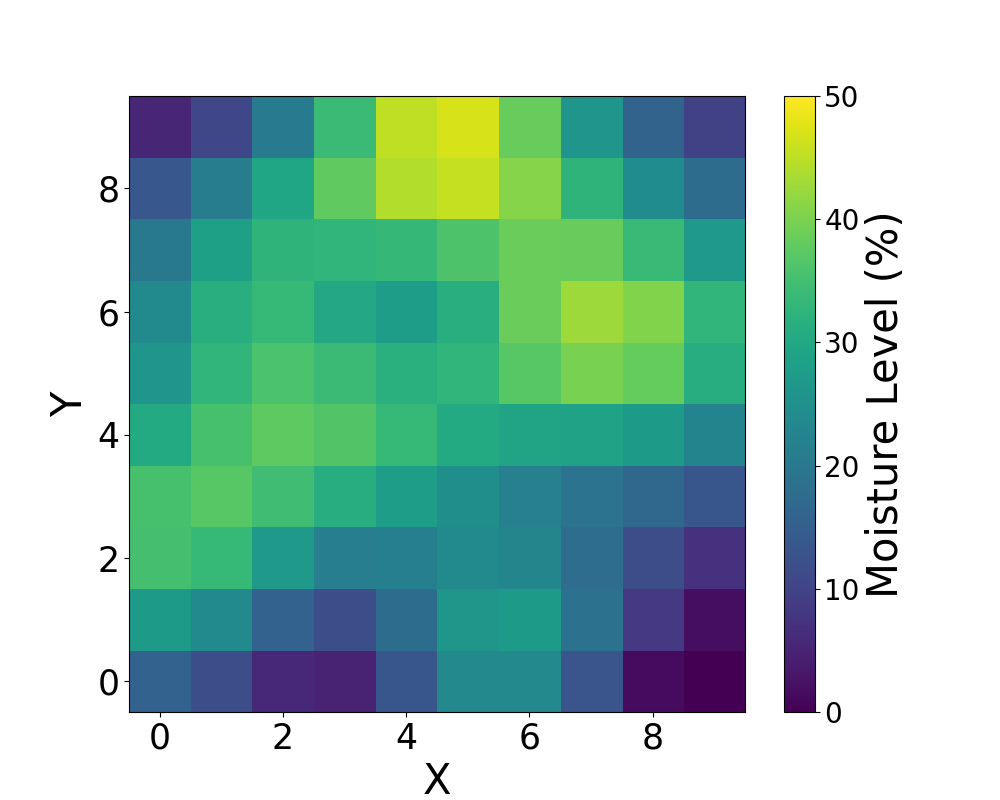}}
    \caption{\normalsize Reconstructed moisture maps from field data. \vspace{-2mm}}
    \label{fig:total_field_reconstruction}
\end{figure}
Moisture maps reconstructed from GT and MoistureMapper  measurements for $A_1$ adaptive sampling and benchmark approaches are shown in Fig. \ref{fig:total_field_reconstruction}. The reconstructed grid for the A1 adaptive sampling and benchmark approaches have RMSE moisture level values of 12.2 and 13.1, respectively.

\section{Conclusion and Future Work}\label{sec:conclusion}

This work presents the design and field deployment results of MoistureMapper, a mobile robot platform designed to measure soil moisture in agricultural fields using TDR sensors. The robot uses an direct push drilling mechanism that enables automated sensor deployment, retraction and moisture measurement. Adaptive sampling algorithms based on a Gaussian Process based modeling were developed and tested in simulation and in field deployments. The computational results show that adaptive sampling strategies outperform a greedy benchmark and allow the system to improve mapping accuracy, reduce uncertainty and travel costs. The proposed robotic platform provides a high resolution moisture mapping solution that can be used by practitioners including farmers and irrigation scientists.

Future work will involve scale-experiments in fields with diverse soil compositions and moisture levels. Adaptive sampling approach based on online estimation of the length parameter in the RBF kernel will be developed to improve moisture mapping accuracy. As discussed in results, due to varying soil composition and moisture levels TDR probes may not be driven fully into soil and instead can result in lifting the robot platform. This can lead to erroneous measurements as well as damage to onboard equipment. We will equip the robot with a force sensing mechanism to auto-stop the DPD system when the probes cannot be drilled any further and use a laser range scanner to measure the penetration depth of the probes. Dakshinamurthy et. al. \cite{dakshinamurthy2024waveform} have shown that TDR electric pulse waveform interpretation can be used to accurately measure soil moisture even in the case of partial insertions. We will build on their method and integrate with automated penetrated probe length measurements to improve accuracy of soil moisture estimation.

\bibliographystyle{IEEEtran}
\bibliography{references}

\end{document}